\documentclass[10pt,twocolumn,letterpaper]{article}

\usepackage[pagenumbers]{cvpr} %

\usepackage{graphicx}
\usepackage{amsmath}
\usepackage{amssymb}
\usepackage{booktabs}
\usepackage{xcolor}
\usepackage{array,makecell}
\usepackage{caption}
\usepackage{graphicx}
\usepackage{wrapfig,lipsum}
\usepackage{enumitem}
\usepackage{amsmath}
\usepackage{bm}
\usepackage{arydshln}
\usepackage[accsupp]{axessibility}

\usepackage[pagebackref,breaklinks,colorlinks]{hyperref}

\usepackage[capitalize]{cleveref}
\crefname{section}{Sec.}{Secs.}
\Crefname{section}{Section}{Sections}
\Crefname{table}{Table}{Tables}
\crefname{table}{Tab.}{Tabs.}

\newcommand{\SMa}{Appendix~\ref{appendix-metrics-and-results}}
\newcommand{\SMb}{Appendix~\ref{appendix-details}~and~\ref{appendix-metrics-and-results}}
\newcommand{\SMc}{Appendix~\ref{appendix-metrics-and-results}}
\newcommand{\SMd}{Appendix~\ref{appendix-details}}
\newcommand{\SMe}{Appendix~\ref{appendix-benchmark}}
\newcommand{\SMf}{Appendix~\ref{appendix-metrics-and-results}}
\newcommand{\AppA}{Section~\ref{section:exp_ablations_analysis} of our main text}

\begin{document}

\title{CODA-Prompt: COntinual Decomposed Attention-based Prompting for Rehearsal-Free Continual Learning}

\author{
\textbf{
James Seale Smith\thanks{Work done partially during internship at MIT-IBM Watson AI Lab.}\,\,\textsuperscript{1,2}
\quad Leonid Karlinsky\textsuperscript{2,4}
\quad Vyshnavi Gutta\textsuperscript{1}
\quad Paola Cascante-Bonilla\textsuperscript{2,3}
}
\\
\textbf{
Donghyun Kim\textsuperscript{2,4}
\quad Assaf Arbelle\textsuperscript{4}
\quad Rameswar Panda\textsuperscript{2,4}
\quad Rogerio Feris\textsuperscript{2,4}
\quad Zsolt Kira\textsuperscript{1}
}
\\
\normalsize
\textsuperscript{1}Georgia Institute of Technology
\quad  \textsuperscript{2}MIT-IBM Watson AI Lab 
\normalsize
\quad \textsuperscript{3}Rice University
\quad \textsuperscript{4}IBM Research
}

\maketitle

\begin{abstract}
    Computer vision models suffer from a phenomenon known as catastrophic forgetting when learning novel concepts from continuously shifting training data. Typical solutions for this continual learning problem require extensive rehearsal of previously seen data, which increases memory costs and may violate data privacy. Recently, the emergence of large-scale pre-trained vision transformer models has enabled prompting approaches as an alternative to data-rehearsal. These approaches rely on a key-query mechanism to generate prompts and have been found to be highly resistant to catastrophic forgetting in the well-established rehearsal-free continual learning setting. However, the key mechanism of these methods is not trained end-to-end with the task sequence. Our experiments show that this leads to a reduction in their plasticity, hence sacrificing new task accuracy, and inability to benefit from expanded parameter capacity. We instead propose to learn a set of prompt components which are assembled with input-conditioned weights to produce input-conditioned prompts, resulting in a novel attention-based end-to-end key-query scheme. Our experiments show that we outperform the current SOTA method DualPrompt on established benchmarks by as much as 4.5\% in average final accuracy. We also outperform the state of art by as much as 4.4\% accuracy on a continual learning benchmark which contains both class-incremental and domain-incremental task shifts, corresponding to many practical settings. Our code is available at \footnotesize{\url{https://github.com/GT-RIPL/CODA-Prompt}}
\end{abstract}

\section{Introduction}
\label{sec:intro}

\begin{figure}[t]
    \centering
    \includegraphics[width=0.46\textwidth]{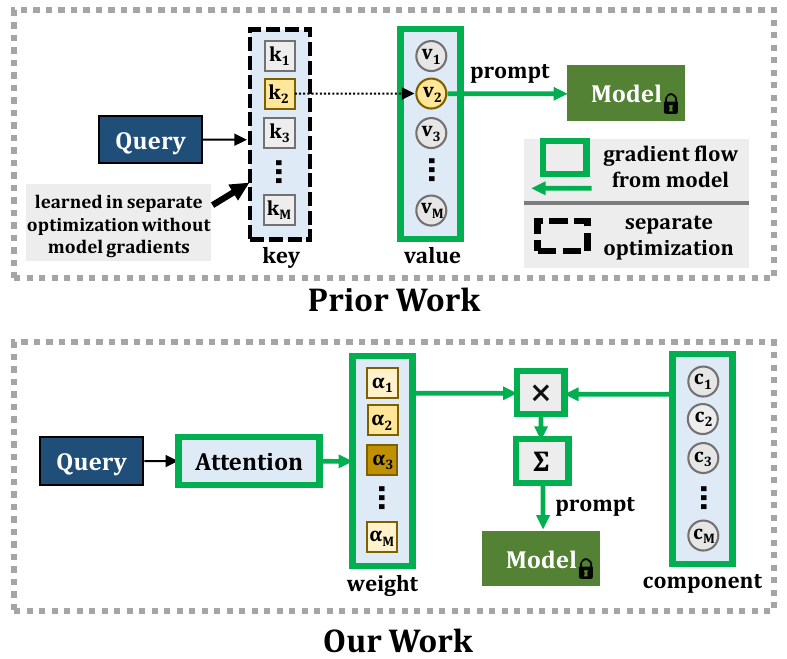}
    \caption{\textbf{Prior work}~\cite{wang2022dualprompt} learns a pool of key-value pairs to select learnable prompts which are inserted into several layers of a pre-trained ViT (the prompting parameters are unique to each layer). \textbf{Our work} introduces a \textbf{decomposed prompt} which consists of learnable prompt \emph{components} that assemble to produce \emph{attention-conditioned} prompts. Unlike prior work, ours is \textbf{optimized in an end-to-end fashion} (denoted with thick, green lines).
    }
    \label{fig:key-idea}
    \vspace{-3mm}
\end{figure}

\vspace{-2mm}
For a computer vision model to succeed in the real world, it must overcome brittle assumptions that the concepts it will encounter after deployment will match those learned \emph{a-priori} during training. The real world is indeed \emph{dynamic} and contains continuously emerging objects and categories. Once deployed,  models will encounter a range of differences from their training sets, requiring us to continuously update them to avoid stale and decaying performance.

One way to update a model is to collect additional training data, combine this new training data with the old training data, and then retrain the model from scratch. While this will guarantee high performance, it is not practical for large-scale applications which may require lengthy training times and aggressive replacement timelines, such as models for self-driving cars or social media platforms. This could incur \emph{high financial~\cite{justus2018predicting} and environmental~\cite{lacoste2019quantifying} costs} if the replacement is frequent. We could instead update the model by only training on the new training data, but this leads to a phenomenon known as \emph{catastrophic forgetting}~\cite{nguyen2019toward} where the model overwrites existing knowledge when learning the new data, a problem known as %
\emph{continual learning}.

The most effective approaches proposed by the continual learning community involve saving~\cite{Rebuffi:2016} or generating~\cite{Shin:2017} a subset of past training data and mixing it with future task data, a strategy referred to as \textbf{rehearsal}. Yet many important applications are unable to store this data because they work with \emph{private user data that cannot be stored} long term.%

In this paper, we instead consider the highly-impactful and well-established setting of \emph{rehearsal-free} continual learning\cite{wang2022learning,wang2022dualprompt,smith2021abd,smith2022closer,li2016learning}\footnote{We focus on continual learning over a single, expanding classification head called \textit{class-incremental continual learning}. This is different from the multi-task continual learning setting, known as \textit{task-incremental} continual learning, where we learn separate classification heads for each task and the task label is provided during inference.\cite{hsu2018re,van2019three}}, limiting our scope to strategies for continual learning which do not store training data. While rehearsal-free approaches have classically under-performed rehearsal-based approaches in this challenging setting by wide-margins~\cite{smith2022closer}, recent \emph{prompt-based approaches}~\cite{wang2022learning,wang2022dualprompt}, leveraging pre-trained Vision transformers (ViT), have had tremendous success and can even outperform state-of-the-art (SOTA) rehearsal-based methods. These prompting approaches boast a strong protection against catastrophic forgetting by learning a small pool of insert-able model embeddings (prompts) rather than modifying vision encoder parameters directly. One drawback of these approaches, however, is that they cannot be optimized in an \emph{end-to-end} fashion as they use a key and query to select a prompt \emph{index} from the pool of prompts, and thus rely on a second, localized optimization to learn the keys because the model gradient cannot backpropagate through the key/query index selection.
Furthermore, these methods reduce forgetting by \emph{sacrificing new task accuracy} (i.e., they lack sufficient \emph{plasticity}). We show that expanding the prompt size does not increase the plasticity, motivating us to grow learning capacity from a new perspective.

We replace the prompt pool with a \emph{decomposed prompt} that consists of a weighted sum of learnable prompt components (Figure~\ref{fig:key-idea}). This decomposition enables higher prompting capacity by expanding in a new dimension (the number of components). Furthermore, this inherently encourages knowledge re-use, as future task prompts will include contributions from prior task components. We also introduce a novel attention-based component-weighting scheme, which allows for our entire method to be \emph{optimized in an end-to-end fashion} unlike the existing works, increasing our plasticity to better learn future tasks. 
We boast significant performance gains on existing rehearsal-free continual learning benchmarks w.r.t. SOTA prompting baselines in addition to a challenging \emph{dual-shift} benchmark containing both incremental concept shifts \emph{and} covariate domain shifts, highlighting our method's impact and generality. Our method improves results even under equivalent parameter counts, but importantly can scale performance by increasing capacity, unlike prior methods which quickly plateau. 
\emph{In summary, we make the following contributions:}
\begin{enumerate}
    \item We introduce a \emph{decomposed} attention-based prompting for rehearsal-free continual learning, characterized by an \emph{expanded learning capacity} compared to existing continual prompting approaches. Importantly, our approach can be optimized in an end-to-end fashion, unlike the prior SOTA.
    \item We establish a new SOTA for rehearsal-free continual learning on the well-established ImageNet-R~\cite{hendrycks2021many,wang2022dualprompt} and CIFAR-100~\cite{krizhevsky2009learning} benchmarks, beating the previous SOTA method DualPrompt~\cite{wang2022dualprompt} by as much as 4.5\%.
    \item We evaluate on a challenging benchmark with dual distribution shifts (semantic \emph{and} covariate) using the ImageNet-R dataset, and again outperform the state of art, highlighting the real-world impact and generality of our approach.
\end{enumerate}

\section{Background and Related Work}
\label{sec:rl}

\noindent
\textbf{Continual Learning}: 
Continual learning approaches can be organized into a few broad categories which are all useful depending on the problem setting and constraints. One group of approaches expands a model's architecture as new tasks are encountered; these are highly effective for applications where a model growing with tasks is practical~\cite{ebrahimi2020adversarial,lee2020neural,lomonaco2017core50,maltoni2019continuous,Rusu:2016}. 
Another approach is to regularize the model with respect to past task knowledge while training the new task. This can either be done by regularizing the model in the  weight space (i.e., penalize changes to model parameters)~\cite{aljundi2017memory,ebrahimi2019uncertainty,kirkpatrick2017overcoming,titsias2019functional,zenke2017continual} or the prediction space (i.e., penalize changes to model predictions)~\cite{Ahn2021ssil,castro2018end,hou2018lifelong,lee2019overcoming,li2016learning}. Regularizing knowledge in the prediction space is done using \textit{knowledge distillation}~\cite{hinton2015distilling} and it has been found to perform better than model regularization-based methods for continual learning when task labels are not given~\cite{lesort2019generative,van2018generative}.

Rehearsal with stored data~\cite{aljundi2019online, aljundi2019gradient,bang2021rainbow,buzzega2020dark,chaudhry2018efficient,chaudhry2019episodic,Gepperth:2017,hayes2018memory,hou2019learning,Kemker:2017,Lopez-Paz:2017,Rebuffi:2016,robins1995catastrophic, rolnick2019experience,von2019continual,wu2019large} or samples from a generative model~\cite{kamra2017deep,kemker2018fear,ostapenko2019learning,Shin:2017,van2020brain} is highly effective when storing training data or training/saving a generative model is possible. Unfortunately, for many machine learning applications long-term storage of training data will violate data privacy, as well as incur a large memory cost. With respect to the generative model, this training process is much more computationally and memory intensive compared to a classification model and additionally may violate data legality concerns because using a generative model increases the chance of memorizing potentially sensitive data~\cite{nagarajan2018theoretical}. This motivates us to work on the important setting of \emph{rehearsal-free} approaches to mitigate catastrophic forgetting.

\noindent
\textbf{Rehearsal-Free Continual Learning}: 
Recent works propose producing images for rehearsal using deep-model inversion~\cite{choi2021dual,gao2022r,smith2021abd,yin2020dreaming}. While these methods perform well compared to generative modeling approaches and simply rehearse from a small number of stored images, model-inversion is a slow process associated with high computational costs in the continual learning setting, and furthermore these methods under-perform rehearsal-based approaches by significant margins~\cite{smith2021abd}. Other works have looked at rehearsal-free continual learning from an online ``streaming" learning perspective using a frozen, pre-trained model~\cite{hayes:2019,lomonaco2020rehearsal}. Because we allow our models to train to convergence on task data (as is common for continual learning~\cite{wu2019large}), our setting is very different. %

\noindent
\textbf{Continual Learning in Vision Transformers}:
Recent work~\cite{brown2020language,li2022technical,yin2019benchmarking} has proven transformers to generalize well to unseen domains. 
For example, one study varies the number of attention heads in Vision transformers and conclude that ViTs offer more robustness to forgetting with respect to CNN-based equivalents~\cite{mirzadeh2022architecture}. Another~\cite{yu2021improving} shows that the vanilla counterparts of ViTs are more prone to forgetting when trained from scratch. Finally, DyTox~\cite{douillard2022dytox} learns a unified model with a parameter-isolation approach and dynamically expands the tokens processed by the last layer to mitigate forgetting. For each task, they learn a new task-specific token per head using task-attention based decoder blocks. However, the above works either rely on exemplars to defy forgetting or they need to train a new transformer model from scratch, a costly operation, hence differing from our objective of exemplar-free CL using pre-trained ViTs.

\noindent
\textbf{Prompting for Continual Learning}: 
Prompt-based approaches for continual learning boast a strong protection against catastrophic forgetting by learning a small number of insert-able model instructions (prompts) rather than modifying encoder parameters directly. The current state-of-the-art approaches
for our setting, DualPrompt~\cite{wang2022dualprompt} and L2P~\cite{wang2022learning}, create a pool of prompts to be selected for insertion into the model, matching input data to prompts \emph{without task id} with a local \emph{clustering-like} optimization. We build upon the foundation of these methods, as discussed later in this paper. We note that the recent S-Prompts~\cite{wang2022sprompt} method is similar to these approaches but designed for \emph{domain-incremental learning} (i.e., learning the same set of classes under covariate distribution task shifts), which is different than the class-incremental continual learning setting of our paper (i.e., learn emerging object classes in new tasks).

\section{Preliminaries}
\label{sec:prelim}

\subsection{Continual Learning}
In our continual learning setting, a model is sequentially shown $N$ tasks corresponding to non-overlapping subsets of semantic object classes. Each class appears in only a single task, and the goal is to incrementally learn to classify new object classes as they are introduced while retaining performance on previously learned classes. To describe our models, we denote $\theta$ as a pre-trained vision encoder and $\phi_{n}$ as our classifier head with logits corresponding to task $n$ classes. In this paper, we deal with the \textit{class-incremental continual learning setting} rather than the \textit{task-incremental continual learning} setting. Class-incremental continual learning is challenging because the learner must support classification across all classes seen up to task $n$~\cite{hsu2018re} (i.e., \textit{no task labels are provided to the learner during inference}). Task-incremental continual learning is a simpler \textit{multi-task} setting where the task labels are given during both training and inference.

\looseness=-1
We also include experiments that contain both \textit{class-incremental \textbf{and} domain-incremental continual learning} \textit{at the same time}. The difference between this setting and the former is that we also inject \emph{covariate distribution shifts} (e.g., task 1 might include real images and clip-art, whereas task 2 might include corrupted photos and artistic paintings). The motivation is to make  continual learning more practical - as in the real world we might encounter these dual types of domain shifts as well.

\subsection{Prompting with Prefix-Tuning}
\label{sec:prelim_b}

In our work, we do not change the technical foundations of \emph{prompting} from the prior state-of-the-art DualPrompt~\cite{wang2022dualprompt}. We instead focus on the \emph{selection and formation} of the prompt (including enabling an end-to-end optimization of all prompting components), and the resulting prompt is used by the vision transformer encoder in an identical fashion to DualPrompt. This allows us to make a fair comparison with respect to our novel contributions.

As done in DualPrompt, we pass prompts to \emph{several} multi-head self-attention (MSA) layers in a pre-trained ViT transformer~\cite{dosovitskiy2020image,vaswani2017attention} and use \emph{prefix-tuning over prompt-tuning}\footnote{We refer the reader to sections 4.2 and 5.4 of the the DualPrompt~\cite{wang2022dualprompt} paper for a discussion on this design choice.}, which prepends prompts to the keys and values of an MSA layer rather than prepending prompts to the input tokens. The layers in which prompting occurs is set as a hyperparameter, and the prompting parameters are unique between layers. We define a prompt parameter as $\bm{p} \in \mathbb{R}^{L_{p} \times D}$ where $L_{p}$ is the prompt length (chosen as a hyperparameter) and $D$ is the embedding dimension ($768$). Consider an MSA layer with input $\bm{h} \in \mathbb{R}^{L \times D}$, and query, key, and values given as $\bm{h}_Q,  \bm{h}_K, \bm{h}_V$, respectively. In the ViT model, $\bm{h}_Q = \bm{h}_K = \bm{h}_V = \bm{h}$. The output of this layer is given as:
\begin{equation}
\begin{aligned}
&\operatorname{ MSA }(\bm{h}_Q, \bm{h}_K, \bm{h}_V) =\operatorname { Concat }\left(\operatorname { h }_{1}, \ldots, \operatorname { h }_{\mathrm{m}}\right) W^{O} \\
&\text { where} \operatorname{h}_{\mathrm{i}} =\operatorname{Attention}\left(\bm{h}_Q W_{i}^{Q}, \bm{h}_K W_{i}^{K}, \bm{h}_V W_{i}^{V}\right)
\label{eq:msa}
\end{aligned}
\end{equation}
where $W^{O}$, $W_{i}^{Q}$, $W_{i}^{K}$, and $W_{i}^{V}$ are projection matrices and $m$ is the number of heads. 
We split our prompt $\bm{p}$ into $\{\bm{p}_K,\bm{p}_V\} \in \mathbb{R}^{\frac{L_{p}}{2} \times D}$ and prepend them to $\bm{h}_K$ and $\bm{h}_V$ with prefix-tuning (P-T) as:
\begin{equation}
f_{P-T}(\bm{p},\bm{h}) = \operatorname{ MSA }(\bm{h}_Q, \left[ \bm{p}_K ;\bm{h}_K \right],\left[\bm{p}_V ;  \bm{h}_V \right]) 
\label{eq:msa_prefix}
\end{equation}
The result is that we now only train a small number of parameter (the prompts) while leaving the rest of the transformer encoder unmodified. The critical question that now remains is \emph{how to select and update prompts in a continuous fashion?} The next subsection will discuss how prior works select and update prompts.

\subsection{L2P and DualPrompt}
L2P~\cite{wang2022learning} and DualPrompt~\cite{wang2022dualprompt} select prompts from a pool using an image-wise prompt query. Specifically, these methods use a key-value pair based query strategy\footnote{We note that the difference between L2P and DualPrompt w.r.t. prompt querying is that the size of the prompt pool in L2P is chosen as a hyperparameter, but in DualPrompt it is equal to the number of tasks and, during training, DualPrompt selects the prompt using task-id (and selects the closest key-query match during inference).} to dynamically select instance-specific prompts from a pool of candidate prompts. Each prompt $\bm{p}_m$ is selected from a pool of learnable keys $ \bm{k}_m \in \mathbb{R}^{D} $ where $M$ is the size of the prompt pool, based on cosine similarity to an input-conditioned query.
Queries are produced as: $q(\bm{x}) \in \mathbb{R}^{D} = \theta(\bm{x})$ where $\theta$ is the pretrained ViT encoder and $x$ is the input image\footnote{The output is read from the class token.}. A query $q(\bm{x})$ is matched to a key $\bm{k}_m$ by selecting the key with the \emph{highest cosine similarity}, denoted as: $\gamma(q(\bm{x}),\bm{k}_m)$. The keys are learned with a separate optimization from the task classification loss by simply maximizing the cosine similarity between each matched $q(\bm{x})$ and $\bm{k}_m$, acting as a \emph{clustering-like} approach. However, this approach does not directly update the key with gradients from the task loss, which is critical in our experiments.
\section{Method}
\label{sec:method}

\begin{figure*}[t]
    \centering
    \centering
    \includegraphics[clip, trim=1.2cm 6cm 3.8cm 3.6cm, width=0.95\textwidth]{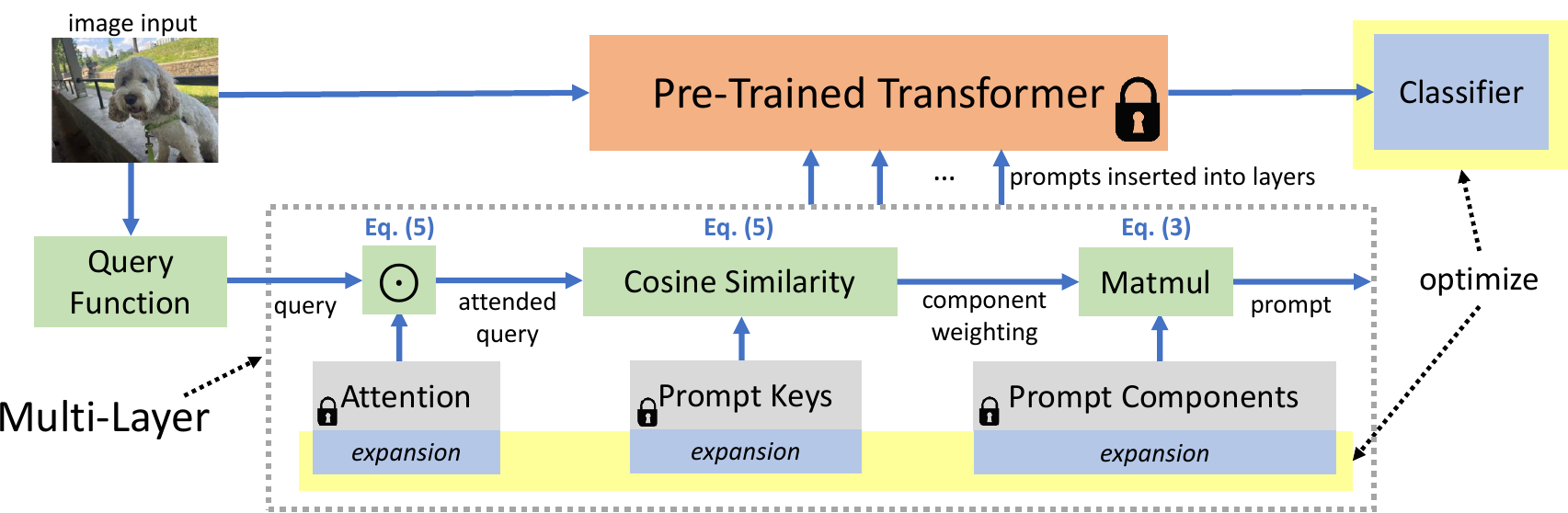}
    \caption{\textbf{Our CODA-Prompt approach to rehearsal-free continual learning}. An input (e.g., image) is passed through a query function (we use the pretrained ViT encoder) and used for a novel \textbf{attention-based prompt-selection scheme} to produce prompts which are passed to multiple layers of a large-scale, pre-trained transformer. Our method is parameterized by an \emph{expanding} set of small prompt components, with each prompt having a corresponding key and attention vector. Both the input and produced prompts are passed through the transformer and sent to the task head. Only the prompting and task parameters are learned which is \emph{parameter efficient} and no training data is stored for replay which is \emph{memory-efficient and privacy preserving.} \emph{Importantly, our prompt selection scheme can be optimized end-to-end, unlike prior works which have a local similarity-based optimization for key-query matching.}
    }
    \label{fig:method}
    \vspace{-3mm}
\end{figure*}

\subsection{Prompt Formation}

While prompting has been shown to perform exceptionally well in terms of mitigating catastrophic forgetting, the existing state-of-the-art approach DualPrompt lacks the ability to expand learning capacity within a given task. Specifically, DualPrompt learns a \emph{single prompt} for each new task - regardless if the task is easy (such as learning a handful of new object classes) or complex (such as learning a large number of new object classes). One might try to increase the length of the prompt, but we show in our experiments (Section~\ref{section:exp_ablations_analysis}) that increasing the prompt length has saturated returns. Intuitively, we instead desire a method where the learning capacity is correlated to the underlying complexity of task data, rather than a single prompt. Additionally, we desire an approach that is end-to-end differentiable, which we conjecture increases the ability to learn a new task with higher accuracy.

We grow our learning capacity by introducing a new axis: \emph{a set of prompt components}. Rather than pick and choose prompts from a pool, we learn a set of prompt components which, via a weighted summation, form a \emph{decomposed}\footnote{Decomposed in the sense that our prompt is a weighted summation of components.} prompt that is passed to the corresponding MSA layer. This allows us to grow our prompting capacity to arbitrary depth and capture the rich underlying complexity of any task while maintaining a fixed prompt length. In addition, prompting in new tasks will inherently reuse the previously acquired knowledge of past tasks rather than initializing a new task prompt from scratch. 

Specifically, we replace learnable prompt parameter $\bm{p}$ with a weighted summation over the prompt components:
\begin{equation}
\bm{p} = \sum_{m} \alpha_m \bm{P_m}
\label{eq:prompt_f}
\end{equation}
where $\bm{P} \in \mathbb{R}^{L_{p} \times D \times M}$ is our set of prompt components, $M$ is the length of our set (i.e., the introduced axis of additional capacity), and $\alpha$ is a weighting vector that determines the contribution of each component which will be discussed in the next subsection.

A distinct difference between our method and L2P/DualPrompt is that \emph{all of our learnable parameters are optimized using the classification loss} rather than splitting into two separate loss optimizations. This allows for better end-to-end learning in our method, and we argue this is a key reason why our method performs better in the benchmarks described in Section~\ref{sec:exp}.

\subsection{Prompt-Component Weighting}

Rather than a pool of prompts, we have a set of prompt components and desire to produce a \emph{weight vector} $\alpha$ given a query $\theta(x)$, rather than a prompt index. We compute the cosine similarity between a query and keys to produce our weighting vector as:
\begin{equation}
\begin{aligned}
\alpha &= \gamma(q(\bm{x}),\bm{K}) \\
&= \{ \gamma(q(\bm{x}),\bm{K}_1),\gamma(q(\bm{x}),\bm{K}_2),\ldots, \gamma(q(\bm{x}),\bm{K}_M)\}
\label{eq:prompt_weight_a}
\end{aligned}
\end{equation}
where $\bm{K} \in \mathbb{R}^{D \times M}$ contains keys corresponding to our prompt components. The intuition here is that the contributing magnitude of each prompt component $\bm{P_m}$ to the final prompt $\bm{p}$ is determined by the similarity between the query $q(\bm{x})$ and corresponding key $\bm{K_m}$.

The prompt-query matching can be thought of as clustering in a high dimension space 
$D$ which is a well-known and difficult problem~\cite{koppen2000curse}. To mitigate this drawback, we introduce another component to our key-query matching: \emph{attention}. Each $\bm{P_m}$ has a corresponding attention vector $\bm{A_m}$ in addition to the key $\bm{K_m}$. This allows the the query to focus on certain features from the high dimensional query $q(\bm{x})$ output - for example, a prompt component used for mammalian face features to classify dogs and cats would be able to focus on relevant query features such as eye placement and skin texture, and furthermore ignore features of unrelated semantics, such as features useful for automobile classification. We use a simple feature-selection \emph{attention} scheme with element-wise multiplication between the query vector and attention vector to create an \emph{attended-query} which is then used for the key-similarity matching. Specifically, our updated approach to producing our weighting vector is:
\begin{equation}
\begin{aligned}
\alpha &= \gamma(q(\bm{x})\odot\bm{A},\bm{K}) \\
&= \{ \gamma(q(\bm{x})\odot\bm{A}_1,\bm{K}_1),\ldots, \gamma(q(\bm{x})\odot\bm{A}_M,\bm{K}_M)\}
\label{eq:prompt_weight_b}
\end{aligned}
\end{equation}
where $\bm{A} \in \mathbb{R}^{D \times M}$ contains learnable parameters (attention vectors) corresponding to our prompt components and $\odot$ is the element-wise product (also known as the Hadamard product). Notice that our attention vectors are learnable feature weightings rather than input-conditioned modules - we found that this fixed representation was less vulnerable to forgetting by its simple design, similar to the our prompt component keys.

\subsection{Expansion \& Orthogonality}
\label{sec:method_expand}

Crucial to mitigating catastrophic forgetting is the avoidance of \emph{overwriting} knowledge acquired in prior tasks. When we visit a new task we freeze our current components and expand the set, only updating the new components. This is visualized in the bottom of Figure~\ref{fig:method} where the existing parameters are \emph{locked} and only the expanded parameters are optimized. Specifically, in task $t$ we learn $\frac{M}{N}$ components where $N$ is the number of tasks and $M$ is a chosen hyperparameter, keeping the previous $\frac{(t-1) \cdot M}{N}$ components \emph{frozen}. This expansion is enabled by our attention-based component-weighting scheme, in that expanding our parameters do \emph{not} alter the computation of weights $\alpha$ corresponding previously learned components. For example, expanding the capacity of a nonlinear model like an MLP module, transformer attention head~\cite{dosovitskiy2020image}, or hyper-network~\cite{von2019continual} would alter the weight outputs of previously learned components. 

While the prior heuristic is focused on reducing catastrophic forgetting (not destroying previously acquired knowledge), we also want to reduce interference between existing and new knowledge. To do this, we add an orthogonality constraint to $\bm{P}$, $\bm{K}$, and $\bm{A}$. The intuition is that orthogonal vectors have less interference with each other. For example, we would want a key and prompt learned in task 2 to not attract task 1 data, otherwise the encoder output for task 1 data will change, resulting in misclassified data. We use an orthogonal initialization and introduce a simple orthogonality penalty loss to our method as:
\begin{equation}
\mathcal{L}_{ortho}(B) = ||B B^\top - I||_2
\label{eq:ortho}
\end{equation}
where $B$ represents any arbitrary matrix.

\subsection{Full Optimization}

Given task classification loss $\mathcal{L}$, our full optimization is:
\begin{equation} \label{eq:full_loss}
\begin{aligned}
    \underset{\bm{P^n}, \bm{K^n}, \bm{A^n}, \phi_{n}}{\operatorname{min}}  &\mathcal{L}(f_\phi(f_{\theta,\bm{P},\bm{K},\bm{A}}(\bm{x})), y) \quad + \\
    \lambda (\mathcal{L}_{ortho} &\left( \bm{P} \right) + \mathcal{L}_{ortho}\left( \bm{K} \right) + \mathcal{L}_{ortho}\left( \bm{A} \right) )
\end{aligned}
\end{equation}
where $\bm{P^n}, \bm{K^n}, \bm{A^n}$ refer to the prompt components and corresponding keys/attention vectors that are unfrozen and trained during task $n$ (see Section~\ref{sec:method_expand}) and $\lambda$ is a hyperparameter balancing the orthogonality loss. Note that we do not update the logits of past task classes, consistent with L2P and DualPrompt. We formally refer to our method as \textbf{CO}ntinual \textbf{D}ecomposed \textbf{A}ttention-based \textbf{P}rompting, or simply \textbf{CODA-P}.

\section{Experiments}
\label{sec:exp}

\begin{table*}[t]
\caption{\textbf{Results (\%) on ImageNet-R}. Results are included for 5 tasks (40 classes per task), 10 tasks (20 classes per task), and 20 tasks (10 classes per task). $A_N$ gives the accuracy averaged over tasks and $F_N$ gives the average forgetting. We report results over 5 trials. %
}
\label{tab:imnet-r_main}
\centering

\begin{tabular}{c||c c||c c||c c} 
\hline 
Tasks & \multicolumn{2}{c||}{5} & \multicolumn{2}{c||}{10} & \multicolumn{2}{c}{20} \\
\hline
\rule{0pt}{10pt} Method & $A_N$ ($\uparrow$) & $F_N$ ($\downarrow$) & $A_N$ ($\uparrow$) & $F_N$ ($\downarrow$) & $A_N$ ($\uparrow$) & $F_N$ ($\downarrow$) \\
\hline
UB  
& $77.13$ & - 
& $77.13$ & - 
& $77.13$ & -   \\ 
\hline
ER (5000)        
& $71.72 \pm 0.71$ & $13.70 \pm 0.26$ 
& $64.43 \pm 1.16$ & $10.30 \pm 0.05$
& $52.43 \pm 0.87$ & $7.70 \pm 0.13$  \\  
FT          
& $18.74 \pm 0.44$ & $41.49 \pm 0.52$
& $10.12 \pm 0.51$ & $25.69 \pm 0.23$
& $4.75 \pm 0.40$ & $16.34 \pm 0.19$  \\
FT++          
& $60.42 \pm 0.87$ & $14.66 \pm 0.24$
& $48.93 \pm 1.15$ & $9.81 \pm 0.31$
& $35.98 \pm 1.38$ & $6.63 \pm 0.11$  \\
LwF.MC       
& $74.56 \pm 0.59$ & $4.98 \pm 0.37$ 
& $66.73 \pm 1.25$ & $3.52 \pm 0.39$
& $54.05 \pm 2.66$ & $2.86 \pm 0.26$  \\
L2P++        
& $70.83 \pm 0.58$ & $3.36 \pm 0.18$
& $69.29 \pm 0.73$ & $2.03 \pm 0.19$ 
& $65.89 \pm 1.30$ & $1.24 \pm 0.14$   \\
Deep L2P++  
& $73.93 \pm 0.37$ & $2.69 \pm 0.10$
& $71.66 \pm 0.64$ & $1.78 \pm 0.16$ 
& $68.42 \pm 1.20$ & $1.12 \pm 0.13$  \\
DualPrompt 
& $73.05 \pm 0.50$ & $\bm{2.64 \pm 0.17}$
& $71.32 \pm 0.62$ & $1.71 \pm 0.24$ 
& $67.87 \pm 1.39$ & $1.07 \pm 0.14$   \\
\hdashline
\textbf{CODA-P-S} 
& $75.19 \pm 0.47$ & $2.65 \pm 0.15$
& $73.93 \pm 0.49$ & $1.60 \pm 0.20$
& $70.53 \pm 1.24$ & $1.00 \pm 0.15$  \\
\textbf{CODA-P}   
& $\bm{76.51 \pm 0.38}$ & $2.99 \pm 0.19$
& $\bm{75.45 \pm 0.56}$ & $\bm{1.64 \pm 0.10}$
& $\bm{72.37 \pm 1.19}$ & $\bm{0.96 \pm 0.15}$  \\ 

\hline
\end{tabular}
\end{table*}

We benchmark our approach with several image datasets in the class-incremental continual learning setting. We implemented baselines which do not store training data for rehearsal: Learning without Forgetting (LwF)~\cite{li2016learning}, Learning to Prompt (L2P)~\cite{wang2022learning}, and DualPrompt~\cite{wang2022dualprompt}. Additionally, we report the upper bound performance (i.e., trained offline) and performance for a neural network trained only on classification loss using the new task training data (we refer to this as FT), and include an improved version of FT which uses the same classifier implementation\footnote{See \SMa{} for additional details} as L2P/DualPrompt (referred to as FT++). We also compare to Experience Replay~\cite{chaudhry2019tiny} to provide additional context to our results. We implement our method and all baselines in PyTorch\cite{paszke2019pytorch} using the ViT-B/16 backbone~\cite{dosovitskiy2020image} pre-trained on ImageNet-1K~\cite{russakovsky2015imagenet}. Our contribution includes faithful PyTorch implementations of the popular prompting baselines L2P and DualPrompt, which were released in JAX\cite{wang2022dualprompt,wang2022learning}. Our implementations of the competing methods actually achieve slight boosts in the performance of DualPrompt in most benchmarks, as well as significant performance boosts in L2P due to the improved prompting type\footnote{As discussed in Section~\ref{sec:prelim_b}, we use pre-fix tuning over prompt-tuning for all implementations as is shown to work best for continual learning~\cite{wang2022dualprompt}.} (which we refer to as L2P++).

DualPrompt uses length 5 prompts in layers 1-2 (referred to as \emph{general} prompts) and length 20 prompts in layers 3-5 (referred to as \emph{task-expert}) prompts.  We insert prompts into the same layers as DualPrompt (layers 1-5) for CODA-P 
and use a prompt length of 8 and 100 prompt components, which were chosen with a hyperparameter sweep on validation data to have the best trade-off between performance and parameter efficiency. Because our approach introduces more learnable parameters compared to DualPrompt, we include a variant of out method \textbf{CODA-P-S} which uses a smaller number of parameters equal to DualPrompt for additional comparison. We show that our method outperforms other methods even under this condition, while still retaining the capability to scale if desired.

We report results on the test dataset, but emphasize that all hyperparameters and design decisions (including for the baselines) were made using validation data (20\% of the training data). Unlike DualPrompt, we run our benchmarks for several different shuffles of the task class order and report the mean and standard deviation of these runs. We do this with a consistent seed (different for each trial) so that results can be directly compared. This is crucial because the results with different shuffling order can be different due to differences in task difficulty and their appearance order. \emph{We include additional implementation details and results in \SMb{}}.

\noindent
\textbf{Evaluation Metrics:} We evaluate methods using (1)
average final accuracy $A_{N}$, or the final accuracy with respect to all past classes averaged over $N$ tasks, and (2) average forgetting~\cite{chaudhry2018efficient,Lopez-Paz:2017,mai2022online} $F_{N}$, or the drop in task performance averaged over $N$ tasks. The reader is referred to \SMc{} for additional details about these metrics. We emphasize that $A_{N}$ is the more important metric and encompasses both method plasticity \emph{and} forgetting, whereas $F_{N}$ simply provides additional context.

\begin{table*}[t]
\caption{\textbf{Results (\%) on CIFAR-100 and DomainNet}. $A_N$ gives the accuracy averaged over tasks and $F_N$ gives the average forgetting. We report results over 5 trials for CIFAR-100 and 3 trials for DomainNet (due to dataset size).}
\begin{subtable}[h]{0.49\textwidth}
\centering
\caption{10-task CIFAR-100 (10 classes per task)}
\label{tab:cifar}
\begin{tabular}{c||c c} 
\hline
\rule{0pt}{10pt} Method  & $A_N$ ($\uparrow$) & $F_N$ ($\downarrow$)  \\
\hline
Upper-Bound & $89.30$ & -  \\  
\hline
ER (5000) & $76.20 \pm 1.04$ & $8.50 \pm 0.37$ \\ 
FT        & $9.92 \pm 0.27$ & $29.21 \pm 0.18$   \\  
FT++      & $49.91 \pm 0.42$ & $12.30 \pm 0.23$   \\ 
LwF.MC    & $64.83 \pm 1.03$ & $5.27 \pm 0.39$   \\ 
L2P++     & $82.50 \pm 1.10$ & $1.75 \pm 0.42$   \\
Deep L2P++ & $84.30 \pm 1.03$ & $\bm{1.53 \pm 0.40}$ \\  
DualPrompt & $83.05 \pm 1.16$ & $1.72 \pm 0.40$  \\
\hdashline
\textbf{CODA-P-S}& $84.59 \pm 0.87$ & $1.76 \pm 0.28$  \\
\textbf{CODA-P}   & $\bm{86.25 \pm 0.74}$ & $1.67 \pm 0.26$   \\

\hline
\end{tabular}

\end{subtable}
\hfill
\begin{subtable}[h]{0.49\textwidth}
\centering
\caption{5-task DomainNet (69 classes per task)}
\label{tab:domainnet}
\begin{tabular}{c||c c} 
\hline
\rule{0pt}{10pt} Method  & $A_N$ ($\uparrow$) & $F_N$ ($\downarrow$) \\
\hline
Upper-Bound  & $79.65$ & -   \\  
\hline
ER (5000)   & $58.32 \pm 0.47$ & $26.25 \pm 0.24$  \\
FT          & $18.00 \pm 0.26$ & $43.55 \pm 0.27$    \\ 
FT++        & $39.28 \pm 0.21$ & $44.39 \pm 0.31$   \\ 
LwF.MC      & $\bm{74.78 \pm 0.43}$ & $5.01 \pm 0.14$ \\
L2P++       & $69.58 \pm 0.39$ & $2.25 \pm 0.08$  \\
Deep L2P++  & $70.54 \pm 0.51$ & $2.05 \pm 0.07$   \\ 
DualPrompt  & $70.73 \pm 0.49$ & $\bm{2.03 \pm 0.22}$  \\
\hdashline
\textbf{CODA-P-S}   & $71.80 \pm 0.57$ & $2.54 \pm 0.10$  \\
\textbf{CODA-P}     & $73.24 \pm 0.59$ & $3.46 \pm 0.09$  \\

\hline
\end{tabular}

\end{subtable}

\end{table*}

\subsection{CODA-P sets the SOTA on existing benchmarks}
We first evaluate our method and state of art on several well-established benchmarks. Table~\ref{tab:imnet-r_main} provides results for ImageNet-R~\cite{hendrycks2021many,wang2022dualprompt} which is composed of 200 object classes with a wide collection of image styles, including cartoon, graffiti, and hard examples from the original ImageNet dataset~\cite{russakovsky2015imagenet}. This benchmark is attractive because the distribution of training data has significant distance to the pre-training data (ImageNet), thus providing a fair and challenging problem setting. In addition to the original 10-task benchmark, we also provide results with a smaller number of large tasks (5-task) and a larger number of small tasks (20 task). We first notice that our reported performance for DualPrompt is a few percentage points higher compared to the original DualPrompt~\cite{wang2022dualprompt} paper. We re-implemented the method from scratch in a different framework (PyTorch~\cite{paszke2019pytorch}) and suspect the difference has to do with our averaging over different class-order shuffling. 
We note that our implementation of L2P (L2P++) performs significantly better than originally reported because we use the same form of prompting as DualPrompt (for fair comparison). Furthermore, our ``Deep L2P ++", which prompts at the same 5 layers as DualPrompt (rather than only at layer 1), actually performs similarly to DualPrompt.

\looseness=-1
Importantly, we see that our method has strong gains in average accuracy across all three task lengths, with as much as \textbf{4.5\% improvement in average accuracy over DualPrompt}. We remind the reader that CODA-P is our proposed method with tuned prompt length and component set size, whereas CODA-P-S is downsized (see \SMd{} for exact details) to exactly match the number of learnable parameters as DualPrompt. We notice that our method often has slightly higher average forgetting compared to DualPrompt. Given that average accuracy is the crucial metric that captures practical performance, we are not worried by the slight uptick in forgetting. In fact, we argue it is very reasonable and reflects the strength of our approach: our method has a larger capacity to learn new tasks, and thus we can sacrifice marginally higher forgetting. Crucially, we see that as the task sequence length grows, the forgetting metric of our method versus DualPrompt begin to converge to a similar value. 

We provide results for experience replay to provide additional context of our results. While the gap between rehearsal with a substantial coreset size (i.e., the buffer size of replay data) and our approach is small for the short task sequence, we see that our method strongly outperforms rehearsal in the longer task sequence.

\looseness=-1 Tables~\ref{tab:cifar}~and~\ref{tab:domainnet} provide results on additional\footnote{Note that the hyperparameters for both DualPrompt and CODA-P were tuned on ImageNet-R, which was intentionally done to evaluate robustness of all methods.} benchmarks ten-task CIFAR-100~\cite{krizhevsky2009learning} and five-task DomainNet~\cite{peng2019moment}. While neither dataset is as impactful as ImageNet-R in terms of challenge and distance from the pre-trained distribution, these tables provide additional context to our method's performance. 
These tables illustrate a similar story to the ImageNet-R benchmarks, with improvements of +3.2\% and +2.5\%, respectively. We do note that, surprisingly, LwF~\cite{li2016learning} slightly outperforms prompting on this benchmark.

\subsection{CODA-P sets the SOTA for a new dual-shift benchmark}

We also evaluate on a challenging \emph{dual-shift} benchmark using the ImageNet-R dataset. Our motivation is to show robustness to two different types of continual distribution shifts: semantic \emph{and} covariate. We accomplish this by randomly selecting image types from the ImageNet-R dataset to include in each task's training data (while keeping the evaluation data unmodified). The reader is refereed to \SMe{} for more details. Results for this benchmark are provided in Table~\ref{tab:imnet-r_domainshift}. Compared to 5-task ImageNet-R where our method improves SOTA by 4.5\% average accuracy, we see a 4.4\% improvement on this 5-task benchmark with similar forgetting margins. Our results indicate that CODA-P better generalizes to real-world type shifts.%

\subsection{Ablations and Additional Analysis}

\label{section:exp_ablations_analysis}

We take a closer look at our method with ablation studies and additional analysis. In Table~\ref{tab:ablations} we show the effect of removing a few key components of our method: attention keys, freezing of past task components, and our orthogonality regularization. We show that removing attention slightly reduces forgetting degrades performance on average accuracy (the more important metric). This makes sense because removing the attention keys makes our query processing more aligned with the L2P/DualPrompt methods which boast low forgetting but lack sufficient learning capacity. We see larger drops when removing freezing and orthogonality. This indicates that these aspects are crucial to our approach. Our intuition is is that, without these, our prompt formation is similar to a shallow MLP module which, unregularized, should suffer from high forgetting.
\begin{table}[t]
\caption{\textbf{Results (\%) on ImageNet-R \emph{with covariate domain shifts}}. Results are included for 5 tasks (40 classes per task). We simulate domain shifts by randomly removing 50\% of the dataset's domains (e.g., clipart, paintings, and cartoon) for the training data of each task (see SM for more details). $A_N$ gives the accuracy averaged over tasks and $F_N$ gives the average forgetting. We report results over 5 trials.}

\centering
\label{tab:imnet-r_domainshift}
\begin{tabular}{c||c c} 
\hline
\rule{0pt}{10pt} Method & $A_N$ ($\uparrow$) & $F_N$ ($\downarrow$) \\
\hline
Upper-Bound   & $77.13$ & $2.66$   \\ 
\hline
ER (5000)    & $67.39 \pm 0.37$ & $11.94 \pm 0.17$  \\
FT           & $17.93 \pm 0.27$ & $37.49 \pm 0.28$ \\
FT++         & $54.51 \pm 0.68$ & $14.41 \pm 0.33$ \\
LwF.MC       & $64.02 \pm 1.55$ & $7.05 \pm 0.27$  \\
L2P++        & $65.08 \pm 0.29$ & $2.79 \pm 0.32$  \\
Deep L2P++   & $65.74 \pm 0.12$ & $2.48 \pm 0.30$   \\ 
DualPrompt   & $66.98 \pm 0.08$ & $\bm{2.21 \pm 0.28}$   \\
\hdashline
\textbf{CODA-P-S}   & $69.73 \pm 0.18$ & $2.35 \pm 0.19$  \\
\textbf{CODA-P}     & $\bm{71.35 \pm 0.08}$ & $2.56 \pm 0.26$  \\

\hline
\end{tabular}
\end{table}

\begin{table}[t]
\caption{\textbf{Ablation Results (\%) on 10-task ImageNet-R}. $A_N$ gives the accuracy averaged over tasks and $F_N$ gives the average forgetting. We report results over 5 trials.}

\centering
\label{tab:ablations}

\begin{tabular}{c||c|c} 
\hline
\rule{0pt}{10pt} Method & $A_N$ ($\uparrow$) & $F_N$ ($\downarrow$) \\
\hline
CODA-P               & $75.45 \pm 0.56$ & $1.64 \pm 0.10$   \\ 
\hline
Ablate Attention     & $74.52 \pm 0.65$ & $1.67 \pm 0.13$   \\  
Ablate Freezing      & $74.60 \pm 0.64$ & $2.29 \pm 0.10$   \\ 
Ablate Orthogonality & $70.66 \pm 0.60$ & $1.74 \pm 0.25$   \\ 

\hline
\end{tabular}
\end{table}

\begin{figure}[t]
    \centering
    \includegraphics[width=0.4\textwidth]{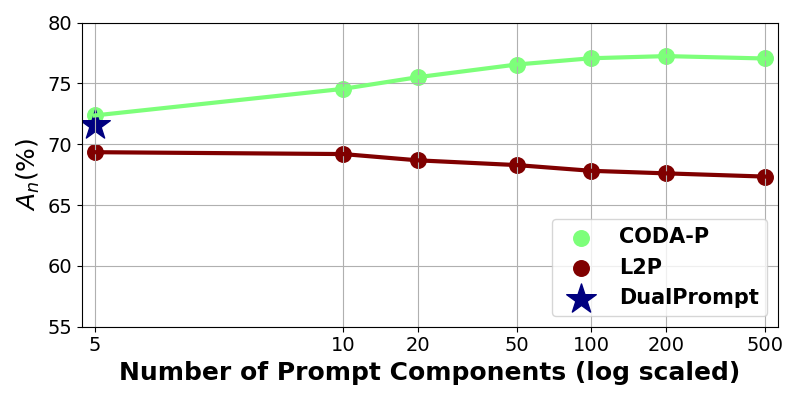}
    \caption{\textbf{Analysis of average accuracy $A_N$ vs prompt components (or pool size)} for CODA-P, L2P, and DualPrompt. The setting is 5 task ImageNet-R, and we report the mean over 3 trials.}
    \label{fig:capacity}
\end{figure}

\begin{figure}[t]
    \centering
    \includegraphics[width=0.4\textwidth]{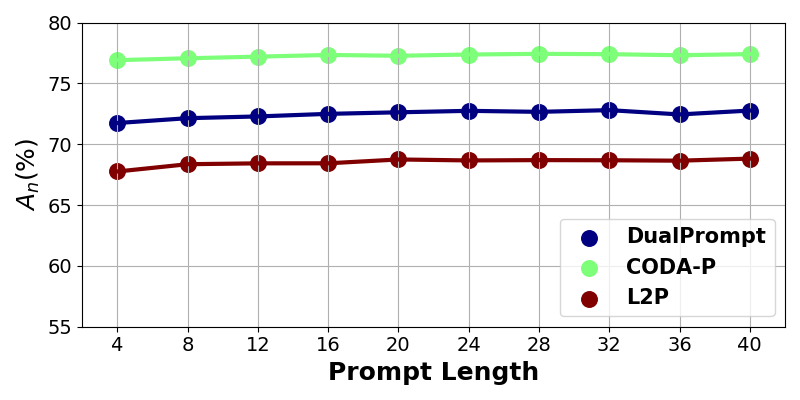}
    \caption{\textbf{Analysis of average accuracy $A_N$ vs prompt length} for CODA-P, L2P, and DualPrompt. The setting is 5 task ImageNet-R, and we report the mean over 3 trials.}
    \label{fig:len}
\end{figure}

\looseness=-1
We also look at our ability to grow in model capacity along our newly introduced prompt component dimension using the 5-task ImageNet-R benchmark (with validation data). We show in Figure~\ref{fig:capacity} that, with a number of prompt components equal to the number of prompts learned in DualPrompt, we achieve higher performance. Importantly, when we increase the number of components, we are able to leverage the scaled parameters for significantly higher performance, finding that our method is closer to upper bound performance than it is to DualPrompt in average accuracy. Because the prompt pool in L2P can be increased to arbitrary size as well, we include results in this analysis as well. However, we show that the L2P method peaks with a pool size equal to twice the task sequence length, and then drops in performance. This reflects that our \emph{prompt components} benefit from scale, whereas the existing \emph{prompt pool} actually suffer. 

\looseness=-1
Finally, we show average accuracy versus prompt \emph{length} in Figure~\ref{fig:len} using the same setting as above. The purpose of this experiment is to emphasize that \emph{accuracy saturates with prompt length}, thus motivating the need to expand in our introduced \emph{component} dimension.
\emph{Additional analysis and hyperparameter sweeps is available in \SMf{}.}

\section{Conclusion}
\label{conclusion}

\looseness=-1
We present \emph{COntinual decomposed attention-based prompting} (\textbf{CODA-Prompt}) for \emph{rehearsal-free continual learning}. Our approach assembles learnable prompt \emph{components} which are then inserted into a pre-trained ViT encoder for image classification. Importantly, CODA-Prompt is optimized in an end-to-end fashion (unlike prior SOTA methods which involve two, separate optimizations). Furthermore, CODA-Prompt can \emph{scale prompting capacity to arbitrary sizes}. We set a new SOTA on both well-established benchmarks and a dual-distribution shift benchmark (containing semantic \emph{and} covariate shifts), highlighting the potential real world impact and generality of our approach.

\section*{Acknowledgements}

This material is based upon work supported by the National Science Foundation under Grant No. 2239292.

{\small
\bibliographystyle{ieee_fullname}
\bibliography{references}
}

\clearpage
\appendix
\section*{Appendix}
\setcounter{figure}{0}
\setcounter{table}{0}
\renewcommand{\thetable}{\Alph{table}}
\renewcommand{\thefigure}{\Alph{figure}}
\renewcommand\thesection{\Alph{section}}
\section{Additional Implementation Details}
\label{appendix-details}

For all methods, we use the Adam~\cite{kingma2014adam} optimizer with $\beta_1 =$ 0.9 and $\beta_1 = $ 0.999, and a batch size of 128 images. We resize all images to 224x224 and normalize them to [0,1]. We train CIFAR-100 and DomainNet for 20 epochs, and ImageNet-R for 50 epochs (chosen to ensure models converge fully for each task). As discussed in the main text, we use the same prompting lengths and locations for L2P~\cite{wang2022learning} and DualPrompt~\cite{wang2022dualprompt} as recommended by the more recent DualPrompt paper. Specifically, for Dualprompt, we use a length 5 prompt in layers 1-2 (referred to as \emph{general} prompts) and length 20 prompts in layers 3-5 (referred to as \emph{task-expert}). For L2P, we use a prompt pool of size 20, total prompt length of size 20, and choose 5 prompts from the pool to use during inference.

As done in DualPrompt~\cite{wang2022dualprompt}, we tuned all additional hyperparameters using 20\% of the training data as validation data. This resulted in using a learning rate of $1e^{-3}$ for all prompting methods (as opposed to $5e^{-3}$ as reported in DualPrompt), and a learning rate of $1e^{-4}$ for all methods which fully fine-tune the model. We searched for learning rates in the values of $\{1e^{-6},5e^{-6},1e^{-5},5e^{-5},1e^{-4},5e^{-4},1e^{-3},5e^{-3},1e^{-2}\}$. We also found that cosine-decaying learning rate outperforms a constant learning rate (which was used in the original DualPrompt implementation). We conjecture that the reduced and decaying learning rate explain the performance boost we obtained on the 10-task ImageNet-R benchmark using our implementations.

For our method, we use a prompt length of 8 and 100 prompt components (and prompt at the same locations as DualPrompt), which were chosen with a hyperparameter sweep on validation data to have the best trade-off between performance and parameter efficiency. We searched for prompt lengths in the range of [4,40] and prompt components in the range of [5,500]. We use $\lambda = $ 0.1 to weight the orthogonality regularization loss, chosen from sweeping across decade values from $1e^{-6}$ up to $1e^2$. As shown in \AppA{}, we see that increasing the prompt length has little effect on our method, whereas increasing the prompt component size has strong returns all the way up to 200 components.

Finally, when implementing classification loss for fine-tuning, L2P, DualPrompt, and CODA-P, we re-use a technique from the official GitHub repo for the DualPrompt and L2P papers~\cite{wang2022learning,wang2022dualprompt} and replace the predictions from past-task logits with negative infinity when training a new task. This results in a softmax prediction of ``0" for these past task classes and prevents gradients from flowing to the linear heads of past task classes. While not discussed in these papers, this technique is \emph{crucial} for performance of these methods, as we confirmed during reproduction. Essentially, the linear layer is highly biased towards new tasks in class-incremental learning in the absence of rehearsal, so this technique prevents the linear head from learning a bias towards new classes over past classes. We note that this bias is a well-known issue~\cite{wu2019large,Ahn2021ssil}.

\section{Additional Results}
\label{appendix-metrics-and-results}

\begin{table*}[t]
\caption{\textbf{Results (\%) on 5-task ImageNet-R (40 classes per task)}. $A_N$ gives the accuracy averaged over tasks, $F_N$ gives the average forgetting, and $N_{param}$ gives the \% of trainable parameters and final parameters w.r.t. the base ViT pre-trained model. We report the mean and standard deviation over 5 trials.}
\vspace{-2mm}
\centering
\label{tab_2:imnet-r_5}
\begin{tabular}{c||c|c|c} 
\hline
\rule{0pt}{10pt} Method  & $A_N$ ($\uparrow$) & $F_N$ ($\downarrow$) & \thead{$N_{param}$ ($\downarrow$) \\ Train/Final}\\
\hline
Upper-Bound  & $77.13$ & - & $100/100$  \\ 
\hline
ER (5000)    & $71.72 \pm 0.71$ & $13.70 \pm 0.26$ & $100/100$  \\ 
FT           & $18.74 \pm 0.44$ & $41.49 \pm 0.52$ & $100/100$  \\
FT++         & $60.42 \pm 0.87$ & $14.66 \pm 0.24$ & $100/100$  \\
LwF.MC       & $74.56 \pm 0.59$ & $4.98 \pm 0.37$ & $100/100$  \\ 
L2P++        & $70.83 \pm 0.58$ & $3.36 \pm 0.18$ & $ 0.7/100.7$  \\
Deep L2P++   & $73.93 \pm 0.37$ & $2.69 \pm 0.10$ & $ 9.6/109.6$  \\ 
DualPrompt   & $73.05 \pm 0.50$ & $\bm{2.64 \pm 0.17}$ & $ 0.5/100.5$  \\
\hdashline
\textbf{CODA-P-S}  & $75.19 \pm 0.47$ & $2.65 \pm 0.15$ & $ 0.7/100.7$  \\
\textbf{CODA-P}    & $\bm{76.51 \pm 0.38}$ & $2.99 \pm 0.19$ & $ 4.6/104.6$  \\

\hline
\end{tabular}

\vspace{9mm}

\caption{\textbf{Results (\%) on 10-task ImageNet-R (20 classes per task)}. $A_N$ gives the accuracy averaged over tasks, $F_N$ gives the average forgetting, and $N_{param}$ gives the \% of trainable parameters and final parameters w.r.t. the base ViT pre-trained model. We report the mean and standard deviation over 5 trials.}
\vspace{-2mm}
\centering
\label{tab_2:imnet-r_10}
\begin{tabular}{c||c|c|c} 
\hline
\rule{0pt}{10pt} Method & $A_N$ ($\uparrow$) & $F_N$ ($\downarrow$) & \thead{$N_{param}$ ($\downarrow$) \\ Train/Final}\\
\hline
Upper-Bound  & $77.13$ & - & $100/100$  \\  
\hline
ER (5000)    & $64.43 \pm 1.16$ & $10.30 \pm 0.05$ & $100/100$  \\ 
FT           & $10.12 \pm 0.51$ & $25.69 \pm 0.23$ & $100/100$  \\ 
FT++         & $48.93 \pm 1.15$ & $9.81 \pm 0.31$ & $100/100$  \\
LwF.MC       & $66.73 \pm 1.25$ & $3.52 \pm 0.39$ & $100/100$ \\
L2P++        & $69.29 \pm 0.73$ & $2.03 \pm 0.19$ & $ 0.7/100.7$  \\
Deep L2P++   & $71.66 \pm 0.64$ & $1.78 \pm 0.16$ & $ 9.6/109.6$  \\  
DualPrompt   & $71.32 \pm 0.62$ & $1.71 \pm 0.24$ & $ 0.8/100.8$  \\
\hdashline
\textbf{CODA-P-S}  & $73.93 \pm 0.49$ & $\bm{1.60 \pm 0.20}$ & $ 0.7/100.7$  \\
\textbf{CODA-P}    & $\bm{75.45 \pm 0.56}$ & $1.64 \pm 0.10$ & $ 4.6/104.6$  \\

\hline
\end{tabular}

\vspace{9mm}
\caption{\textbf{Results (\%) on 20-task ImageNet-R (10 classes per task)}. $A_N$ gives the accuracy averaged over tasks, $F_N$ gives the average forgetting, and $N_{param}$ gives the \% of trainable parameters and final parameters w.r.t. the base ViT pre-trained model. We report the mean and standard deviation over 5 trials.}
\vspace{-2mm}
\centering
\label{tab_2:imnet-r_20}
\begin{tabular}{c||c|c|c} 
\hline
\rule{0pt}{10pt} Method & $A_N$ ($\uparrow$) & $F_N$ ($\downarrow$) & \thead{$N_{param}$ ($\downarrow$) \\ Train/Final}\\
\hline
Upper-Bound  & $77.13$ & - & $100/100$  \\ 
\hline
ER (5000)    & $52.43 \pm 0.87$ & $7.70 \pm 0.13$ & $100/100$  \\ 
FT           & $4.75 \pm 0.40$ & $16.34 \pm 0.19$ & $100/100$  \\ 
FT++         & $35.98 \pm 1.38$ & $6.63 \pm 0.11$ & $100/100$  \\
LwF.MC       & $54.05 \pm 2.66$ & $2.86 \pm 0.26$ & $100/100$  \\
L2P++        & $65.89 \pm 1.30$ & $1.24 \pm 0.14$ & $ 0.7/100.7$  \\
Deep L2P++   & $68.42 \pm 1.20$ & $1.12 \pm 0.13$ & $ 9.6/109.6$  \\ 
DualPrompt   & $67.87 \pm 1.39$ & $1.07 \pm 0.14$ & $ 1.3/101.3$  \\
\hdashline
\textbf{CODA-P-S}  & $70.53 \pm 1.24$ & $1.00 \pm 0.15$ & $ 0.7/100.7$  \\
\textbf{CODA-P}    & $\bm{72.37 \pm 1.19}$ & $\bm{0.96 \pm 0.15}$ & $ 4.6/104.6$  \\

\hline
\end{tabular}
\end{table*}

\begin{table*}[t]
\caption{\textbf{Results (\%) on 10-task CIFAR-100 (10 classes per task)}. $A_N$ gives the accuracy averaged over tasks, $F_N$ gives the average forgetting, and $N_{param}$ gives the \% of trainable parameters and final parameters w.r.t. the base ViT pre-trained model. We report the mean and standard deviation over 5 trials.}
\vspace{-2mm}
\centering
\label{tab_2:cifar}
\begin{tabular}{c||c|c|c} 
\hline
\rule{0pt}{10pt} Method & $A_N$ ($\uparrow$) & $F_N$ ($\downarrow$) & \thead{$N_{param}$ ($\downarrow$) \\ Train/Final}\\
\hline
Upper-Bound  & $89.30$ & - & $100/100$  \\  
\hline
ER (5000)    & $76.20 \pm 1.04$ & $8.50 \pm 0.37$ & $100/100$  \\ 
FT           & $9.92 \pm 0.27$ & $29.21 \pm 0.18$ & $100/100$  \\  
FT++         & $49.91 \pm 0.42$ & $12.30 \pm 0.23$ & $100/100$  \\  
LwF.MC       & $64.83 \pm 1.03$ & $5.27 \pm 0.39$ & $100/100$  \\ 
L2P++        & $82.50 \pm 1.10$ & $1.75 \pm 0.42$ & $ 0.7/100.7$  \\
Deep L2P++   & $84.30 \pm 1.03$ & $\bm{1.53 \pm 0.40}$ & $ 9.5/109.5$  \\  
DualPrompt   & $83.05 \pm 1.16$ & $1.72 \pm 0.40$ & $ 0.7/100.7$  \\
\hdashline
\textbf{CODA-P-S}  & $84.59 \pm 0.87$ & $1.76 \pm 0.28$ & $ 0.6/100.6$  \\
\textbf{CODA-P}    & $\bm{86.25 \pm 0.74}$ & $1.67 \pm 0.26$ & $ 4.6/104.6$  \\

\hline
\end{tabular}

\vspace{9mm}

\caption{\textbf{Results (\%) on 5-task DomainNet (69 classes per task)}. $A_N$ gives the accuracy averaged over tasks, $F_N$ gives the average forgetting, and $N_{param}$ gives the \% of trainable parameters and final parameters w.r.t. the base ViT pre-trained model. We report the mean and standard deviation over 3 trials.}
\vspace{-2mm}
\centering
\label{tab_2:domainnet}
\begin{tabular}{c||c|c|c} 
\hline
\rule{0pt}{10pt} Method  & $A_N$ ($\uparrow$) & $F_N$ ($\downarrow$) & \thead{$N_{param}$ ($\downarrow$) \\ Train/Final}\\
\hline
Upper-Bound  & $79.65$ & - & $100/100$  \\ 
\hline
ER (5000)    & $58.32 \pm 0.47$ & $26.25 \pm 0.24$ & $100/100$  \\
FT           & $18.00 \pm 0.26$ & $43.55 \pm 0.27$ & $100/100$  \\ 
FT++         & $39.28 \pm 0.21$ & $44.39 \pm 0.31$ & $100/100$ \\
LwF.MC       & $\bm{74.78 \pm 0.43}$ & $5.01 \pm 0.14$ & $100/100$  \\
L2P++        & $69.58 \pm 0.39$ & $2.25 \pm 0.08$ & $ 0.9/100.9$  \\
Deep L2P++   & $70.54 \pm 0.51$ & $2.05 \pm 0.07$ & $ 9.7/109.7$  \\  
DualPrompt   & $70.73 \pm 0.49$ & $\bm{2.03 \pm 0.22}$ & $ 0.6/100.6$  \\
\hdashline
\textbf{CODA-P-S}    & $71.80 \pm 0.57$ & $2.54 \pm 0.10$ & $ 0.6/100.6$  \\
\textbf{CODA-P}      & $73.24 \pm 0.59$ & $3.46 \pm 0.09$ & $ 4.8/104.8$  \\

\hline
\end{tabular}

\vspace{9mm}

\caption{\textbf{Results (\%) on ImageNet-R \emph{with covariate domain shifts}}. Results are included for 5 tasks (40 classes per task). We simulate domain shifts by randomly removing 50\% of the dataset's domains (e.g., clipart, paintings, and cartoon) for the training data of each task (see SM for more details). $A_N$ gives the accuracy averaged over tasks, $F_N$ gives the average forgetting, and $N_{param}$ gives the \% of trainable parameters and final parameters w.r.t. the base ViT pre-trained model. We report the mean and standard deviation over 5 trials.}
\vspace{-2mm}
\centering
\label{tab_2:imnet-r_domainshift}
\begin{tabular}{c||c|c|c} 
\hline
\rule{0pt}{10pt} Method & $A_N$ ($\uparrow$) & $F_N$ ($\downarrow$) & \thead{$N_{param}$ ($\downarrow$) \\ Train/Final}\\
\hline
Upper-Bound  & $77.13$ & - & $100/100$  \\  
\hline
ER (5000)    & $67.39 \pm 0.37$ & $11.94 \pm 0.17$ & $100/100$  \\  
FT           & $17.93 \pm 0.27$ & $37.49 \pm 0.28$ & $100/100$  \\
FT++         & $54.51 \pm 0.68$ & $14.41 \pm 0.33$ & $100/100$  \\
LwF.MC       & $64.02 \pm 1.55$ & $7.05 \pm 0.27$ & $100/100$  \\
L2P++        & $65.08 \pm 0.29$ & $2.79 \pm 0.32$ & $ 0.7/100.7$  \\
Deep L2P++   & $65.74 \pm 0.12$ & $2.48 \pm 0.30$ & $ 9.6/109.6$ \\  
DualPrompt   & $66.98 \pm 0.08$ & $\bm{2.21 \pm 0.28}$ & $ 0.8/100.8$  \\
\hdashline
\textbf{CODA-P-S}  & $69.73 \pm 0.18$ & $2.35 \pm 0.19$ & $ 0.7/100.7$  \\ 
\textbf{CODA-P}    & $\bm{71.35 \pm 0.08}$ & $2.56 \pm 0.26$ & $ 4.6/104.6$  \\

\hline
\end{tabular}
\end{table*}

We report extended results, including standard deviations and additional parameters trained, for all benchmarks in Tables \ref{tab_2:imnet-r_5} (5-task ImageNet-R~\cite{hendrycks2021many,wang2022dualprompt}), \ref{tab_2:imnet-r_10} (10-task ImageNet-R), \ref{tab_2:imnet-r_20} (20-task Imagenet-R), \ref{tab_2:cifar} (10-task CIFAR-100~\cite{krizhevsky2009learning}), \ref{tab_2:domainnet} (5-task DomainNet~\cite{peng2019moment}), and \ref{tab_2:imnet-r_5} (Dual-Shift ImageNet-R).

We evaluate methods using (1) final average accuracy $A_{N}$, or the final model's test accuracy averaged over all $N$ tasks, and (2) average forgetting~\cite{chaudhry2018efficient,Lopez-Paz:2017,mai2022online} $F_{N}$, or the drop in task performance averaged over $N$ tasks. The reader is referred to Appendix C of Wang \emph{et al.}~\cite{wang2022dualprompt} for the formal metric definitions.
We emphasize that $A_{N}$ is the more important metric and encompasses both method plasticity \emph{and} forgetting, 
whereas $F_{N}$ provides additional context subject to the model's plasticity (i.e., a lower $F_{N}$ value \emph{and} a lower $A_{N}$ value would indicate that the model's lower forgetting results from its reduced adaptivity to new tasks, which is an undesirable trait).

For each result, we calculate the mean and standard deviation over separate runs. Each run contains different shuffles of the class order; specifically, we shuffle the classes using a random seed that is set for each ``trial run" - and form the tasks using this class shuffle. \emph{Importantly, the class order and all randomized seeds, including model initialization, are consistent between different methods in the same ``trial run".}

We additionally report the number of parameters \emph{trained} (i.e., unlocked during training a task) as well as the \emph{total} number of parameters in the final model. These are reported in \% of the backbone model for easy comparison. Importantly, we design CODA-P-S to have fewer parameters than DualPrompt in the 10-task ImageNet-R setting (our main experiment setting). As we change the number of tasks in ImageNet-R, the number of total parameters for DualPrompt changes because the pool size is set as equal to the number of total tasks by definition (unlike ours, which is set as a hyper-parameter, allowing us to increase or decrease the number of trainable parameters to accommodate the underlying complexity of the task.)

\section{ImageNet-R Dual-Shift Benchmark}
\label{appendix-benchmark}

Our motivation for the challenging \emph{dual-shift} ImageNet-R~\cite{hendrycks2021many,wang2022dualprompt} benchmark is to show robustness to two different types of continual distribution shifts: semantic \emph{and} covariate. Specifically, there are 15 image types in the ImageNet-R dataset: \emph{`art', `cartoon', `deviantart', `embroidery', `graffiti', `graphic', `misc', `origami', `painting', `sculpture', `sketch', `sticker', `tattoo', `toy', `videogame'}. We divide the dataset into 5 tasks of 40 classes each, and within each task we randomly remove 8 of the domain types from the training data. The task becomes more challenging because we now have image type domain shifts injected into the continual learning task sequence, in addition to the already-present class-incremental shifts.

\end{document}